\documentclass[letterpaper]{article}
\usepackage[preprint]{aaai2027}
\usepackage{amsmath}
\usepackage{amssymb}
\usepackage{booktabs}
\usepackage{array}
\usepackage{graphicx}
\usepackage{caption}
\usepackage{newfloat}
\DeclareFloatingEnvironment[name=Algorithm,placement=t]{algorithm}
\usepackage{tikz}
\usetikzlibrary{arrows.meta,positioning,fit,backgrounds,calc}
\usepackage[hyphens]{url}
\usepackage{natbib}
\title{UC-Search: Risk-Aware Test-Time Search for Delayed Constrained Time-Series Control}
\author{
Xibai Wang
}
\affiliations{
NeuroQuant Labs Limited\\
Hong Kong SAR, China\\
xibai.wang@neuroquantlabs.com
}

\begin{document}
\maketitle

\begin{abstract}
Time-series deployments often need delayed feasible decisions, not only accurate forecasts. UC-Search is a trace-only retained-search layer for delayed constrained control: a frozen backbone emits forecasts or action scores, a hard-feasibility automaton rolls paths forward, and bounded search returns the first action of a feasible trajectory. The main claim is conditional: retained lookahead can improve delayed constrained decisions only when delayed feasible-set coupling, retained-prefix premises, and fail-closed release certificates hold. The promoted public endpoint is Phase128 certified M4 expanded40: validation selects Certificate-Constrained Retained Pareto Beam with $\lambda=0.25$, the held-out test has certificate/risk-active rates $1.0000/0.9642$, and the weakest family remains above the unchanged $0.95$ gate at $0.9516$ on M4Weekly. The author-defined public $9$-family suite remains an uncertified stress-test boundary. The paper reports a trace-only mechanism, one certified public endpoint, failed-route certificates, and deployment boundaries rather than a universal risk-control theorem.
\end{abstract}

\section{Introduction}

Time-series learning has advanced through PatchTST \cite{nie2023patchtst}, TimesNet \cite{wu2023timesnet}, Temporal Fusion Transformers \cite{lim2021tft}, DLinear \cite{zeng2023linear}, iTransformer \cite{liu2024itransformer}, and foundation-style forecasters such as TimesFM, Chronos, MOMENT, Lag-Llama, and Moirai \cite{das2024timesfm,ansari2024chronos,goswami2024moment,rasul2023lagllama,woo2024moirai}. These models primarily optimize forecasting or representation objectives, while many deployments need decisions: trade, alarm, intervene, defer, or allocate.

Reinforcement learning methods such as DQN \cite{mnih2015dqn} and PPO \cite{schulman2017ppo} address sequential decisions, but they are training-centric and execute a learned policy at test time. TENT \cite{wang2021tent} adapts parameters at inference, and ReAct or Tree of Thoughts add structured test-time computation for language tasks \cite{yao2023react,yao2023tot}; neither directly gives a lightweight search layer for numeric time-series decisions.

\begin{figure}[!t]
\centering
\begin{tikzpicture}[
font=\sffamily\footnotesize,
node distance=4.5mm and 7.5mm,
box/.style={draw, rounded corners=1.7pt, align=center, minimum height=7.0mm, text width=.30\columnwidth, inner sep=2.0pt, line width=.5pt},
blue/.style={box, fill=blue!5, draw=blue!60!black},
green/.style={box, fill=green!7, draw=green!45!black},
orange/.style={box, fill=orange!9, draw=orange!70!black},
graybox/.style={box, fill=black!3, draw=black!55},
arrow/.style={-{Latex[length=1.55mm]}, line width=.5pt},
edge label/.style={fill=white, inner sep=.7pt, font=\sffamily\footnotesize}
]
\node[blue] (trace) at (0,0) {frozen trace\\$\hat q(s,a), u(s,a)$};
\node[green, right=of trace] (auto) {hard feasibility\\$F(s),T(s,a)$};
\node[orange, below=of trace] (search) {retained search\\Beam / MCTS};
\node[orange, right=of search] (score) {path score\\$\hat R(p)-\lambda U(p)$};
\node[green, below=of search] (gate) {certificate gate\\$\mathcal R_{\rm cert}$};
\node[graybox, right=of gate] (fallback) {fallback +\\no-go audit};
\node[green, below=of gate] (release) {released\\first action};
\node[orange, right=of release] (claim) {claim boundary\\promote / boundary};

\draw[arrow] (trace) -- (auto);
\draw[arrow] (trace) -- (search);
\draw[arrow] (auto) -- (score);
\draw[arrow] (search) -- (score);
\draw[arrow] (score.south west) -- (gate.north east);
\draw[arrow] (gate) -- node[edge label, left=1.2pt, pos=.54]{pass} (release);
\draw[arrow] (gate) -- node[edge label, above=1.0pt, pos=.48]{fail} (fallback);
\draw[arrow] (release) -- (claim);
\draw[arrow] (fallback) -- (claim);
\end{tikzpicture}
\caption{UC-Search dispatch architecture: offline profiles and hard feasibility feed an online retained-search spine; certificate failures route to fallback and no-go audit evidence rather than an uncertified release.}
\label{fig:architecture}
\end{figure}
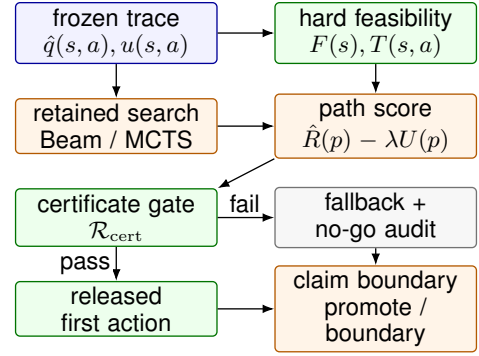

UC-Search studies trace-only test-time decision search guided by predictive uncertainty. Unlike finite-action forecast-MPC, scenario MPC, or risk-sensitive MCTS variants that reason over planner-native models, scenarios, or simulator rollouts, UC-Search consumes forecast/action traces plus a hard-feasibility automaton and returns a first action without retraining. The novelty claim is not beam search or receding-horizon execution itself, but the trace-only interface, validation-selected first-action contract, retained-search certificate, and release-gated claim separation. The empirical story is deliberately split: delayed inventory is certificate-audited retained-search evidence without dominance promotion; M4 lead-time inventory is standard-domain delayed-search corroboration with a $\lambda=0$/MPPI boundary; the public multi-family delayed-control benchmark is an author-defined stress-test boundary documented in a benchmark card, not a community-standard benchmark; retained-prefix theory is a mechanism and failure certificate, not a universal explanation of all positive rows.

The intended contributions are:
\begin{enumerate}
    \item a forecast-trace + hard-feasibility automaton + uncertainty-guided first-action search formulation for delayed constrained time-series control problems;
    \item UC-Beam as the primary bounded-search policy, UC-MCTS as a hard-feasible UCT-style diagnostic variant, and a certificate-constrained retained Pareto Beam route that treats a fail-closed certificate constraint as part of the decision rule rather than only a post-hoc diagnostic;
    \item myopic-collapse, delayed feasible-set separation, retained-proxy regret, retained-prefix margin, and auditable certificate results for when bounded retained lookahead can improve first-action selection;
    \item a reproducible implementation with a benchmark card, primary evidence, standard corroboration, mechanism evidence, boundary diagnostics, deployment boundaries, and release-gated reproducibility artifacts separated by an explicit claim taxonomy;
    \item an explicit closest-work contrast showing where the current formulation differs from, and remains weaker than, finite-action forecast-MPC, risk-sensitive MCTS, decision-focused learning, and model-based RL.
\end{enumerate}

\section{Related Work}

\paragraph{Time-series forecasting.}
PatchTST, TimesNet, TFT, DLinear, and iTransformer improve long-horizon or general time-series forecasting \cite{nie2023patchtst,wu2023timesnet,lim2021tft,zeng2023linear,liu2024itransformer}; DeepAR and foundation-style forecasters provide distributional or zero-shot forecasts \cite{salinas2017deepar,das2024timesfm,ansari2024chronos,goswami2024moment,rasul2023lagllama,woo2024moirai}. These works provide candidate backbones but not a test-time search layer over downstream decision trajectories.

\paragraph{Uncertainty estimation and constrained control.}
Kendall and Gal distinguish aleatoric from epistemic uncertainty \cite{kendall2017uncertainties}; deep ensembles and adaptive conformal prediction provide scalable predictive-uncertainty estimates for learning systems and time series \cite{lakshminarayanan2017deep,zaffran2022adaptive}. Stochastic MPC uses uncertainty to protect constraint satisfaction rather than only to calibrate forecasts \cite{mesbah2016smpc}. UC-Search is closer to this receding-horizon view than to pure forecast scoring, but assumes only forecast/action-score traces and a hard feasibility automaton.

\paragraph{Decision learning, MPC, and search.}
Deep RL learns policies or value functions \cite{mnih2015dqn,schulman2017ppo}; PETS, MBPO, and MuZero add learned dynamics, uncertainty, sampling, or planning \cite{chua2018pets,janner2019mbpo,schrittwieser2020muzero}. Classical MPC optimizes a horizon and executes the first action \cite{rawlings2017mpc}; finite-action forecast-MPC and scenario MPC would also solve a first-action horizon problem if a planner-facing transition, cost, and scenario model are available. CEM and MPPI are common derivative-free model-predictive optimizers \cite{rubinstein1999cem,williams2017mppi,chua2018pets}. MCTS/UCT and AlphaGo show neural tree-search value \cite{kocsis2006uct,silver2016alphago}; risk-sensitive MCTS variants can propagate risk objectives inside simulator-backed trees. Closest to UC-Search, UA-MCTS, Wasserstein MCTS, Bayes-adaptive MCTS, and EMCTS study uncertainty-aware planning in stronger simulator/model-state frameworks \cite{kohankhaki2024uamcts,dam2025wasserstein,chen2026bayesadaptive,oren2025emcts}. The contribution here is narrower: a trace-only wrapper for finite-action constrained delayed control, not a general MCTS regret, broad MPC/MBRL improvement, or proof that MPC-style methods are beaten broadly; CEM/MPPI are validation-selected same-task stochastic optimizer controls.

\paragraph{Decision-focused and financial forecasting.}
Predict-then-optimize and decision-focused learning train predictors against downstream decision losses \cite{elmachtoub2020spo,mandi2024dflsurvey}. UC-Search instead asks whether a fixed trace plus feasibility layer can improve delayed decisions under matched compute; the decision-focused baseline is therefore conceptually upstream and remains a boundary rather than a replacement for the reported frozen-trace contract. In finance, FI-2010 is a public LOB mid-price benchmark \cite{ntakaris2018fi2010}, DeepLOB is a standard LOB forecasting model, and BDLOB shows that Bayesian uncertainty can improve trading decisions \cite{zhang2019deeplob,zhang2018bdlob}. These works motivate FI-2010 boundary evaluation, while UC-Search studies uncertainty as a search-expansion and path-selection signal rather than a full BDLOB-style trading system.

\paragraph{Inventory control and forecasting benchmarks.}
Periodic-review lost-sales inventory has classical base-stock and $(s,S)$ controls \cite{scarf1960ss,zipkin2000inventory}. The M4 competition provides a large public forecasting benchmark \cite{makridakis2020m4}. The M4 lead-time suite below uses raw demand and classical controls to test a standard-domain delayed decision setting; because its selected UC configuration has $\lambda=0$, it is delayed-search evidence, not evidence that uncertainty penalties are always needed.

\begin{table}[t]
\centering
\footnotesize
\setlength{\tabcolsep}{1.2pt}
\begin{tabular}{@{}p{.20\columnwidth}p{.20\columnwidth}p{.13\columnwidth}p{.12\columnwidth}p{.23\columnwidth}@{}}
\toprule
Method & Planner input & State & Learns? & Boundary \\
\midrule
Forecast-MPC & model and scenarios & plant & no & needs transition model \\
CEM/MPPI & simulator & rollout & no & same-budget control \\
Risk-MCTS & simulator tree & planner & no & stronger rollout state \\
DFL & train data & pred. & yes & changes predictor \\
UC-Search & trace + automaton & wrapper & no & frozen trace contract \\
\bottomrule
\end{tabular}
\caption{What is new? UC-Search combines a trace-only hard-feasibility automaton, validation-selected first-action contract, and retained-prefix certificate; its claim is narrower than finite-action forecast-MPC, risk-sensitive MCTS, model-based RL, and decision-focused learning.}
\label{tab:contrast}
\end{table}

\section{Problem Definition}

At time $t$, the decision maker observes a time-series window
\begin{equation}
X_t = \{x_{t-k}, \ldots, x_t\},
\end{equation}
and chooses an action from a finite action space
\begin{equation}
A = \{a_1, \ldots, a_m\}.
\end{equation}
A backbone model produces action-value estimates, forecasts, or policy scores. UC-Search constructs a bounded-depth tree of candidate action paths and returns the first action of the best path:
\begin{equation}
a_t^\star = \operatorname{first}\left(
\arg\max_{p \in \mathcal{T}_t}
 \hat{R}(p) - \lambda U(p)
\right),
\end{equation}
where $\hat{R}(p)$ is estimated reward, $U(p)$ is propagated uncertainty, and $\lambda$ controls risk sensitivity.

\begin{table}[t]
\centering
\footnotesize
\setlength{\tabcolsep}{2.4pt}
\begin{tabular}{@{}ll@{}}
\toprule
Symbol & Meaning \\
\midrule
$X_t$ & observed time-series window at decision time $t$ \\
$A, a$ & finite action set and one candidate action \\
$p$ & candidate action path; $\operatorname{first}(p)$ is executed \\
$\mathcal{T}_t$ & bounded retained search tree at time $t$ \\
$\hat q(s,a)$ & backbone reward/action score for local state $s$ \\
$u(s,a)$ & local predictive uncertainty used during expansion \\
$C(s,a)$ & top-$k$ candidate expansion score \\
$\hat R(p), U(p)$ & path reward estimate and propagated uncertainty \\
$S(p)$ & risk-adjusted retained-path score \\
$\lambda,\lambda_c$ & path risk penalty and local pruning penalty \\
$k,B,d$ & candidate cap, beam width, and search depth \\
$F(s),T(s,a)$ & feasible actions and planner transition callback \\
$\mathcal{R}_{\mathrm{cert}}$ & selective certified-release set \\
$\epsilon_{\mathrm{cert}}$ & executable selective regret/certificate slack \\
\bottomrule
\end{tabular}
\caption{Key notation for the UC-Search formulation and certificate gate.}
\label{tab:notation}
\end{table}

\section{Method}

\subsection{Uncertainty Estimation}

For an ensemble of predictions $\{\hat{y}^{(j)}\}_{j=1}^J$, UC-Search estimates the predictive mean and epistemic uncertainty as
\begin{equation}
\mu = \frac{1}{J}\sum_{j=1}^J \hat{y}^{(j)}, \quad
U_{\mathrm{epi}} = \operatorname{Var}_{j}(\hat{y}^{(j)}).
\end{equation}
If a heteroscedastic head predicts observation variance, the aleatoric component is
\begin{equation}
U_{\mathrm{ale}} = \frac{1}{J}\sum_{j=1}^J \sigma_j^2.
\end{equation}
For a path $p = (a_t,\ldots,a_{t+d})$, uncertainty is propagated as
\begin{equation}
U(p)=\sum_{\ell=0}^{d}\gamma^\ell
\left(U_{\mathrm{epi}}(t+\ell,a_{t+\ell})+
U_{\mathrm{ale}}(t+\ell,a_{t+\ell})\right).
\end{equation}

\subsection{Candidate Expansion}

At each node, a candidate generator selects the top-$k$ actions under an uncertainty-adjusted local score
\begin{equation}
C(s,a)=\hat{q}(s,a)-\lambda_c u(s,a)+\alpha u(s,a),
\end{equation}
where $\hat{q}$ is the backbone action score, $u$ is local predictive uncertainty, $\lambda_c$ controls risk pruning, and $\alpha$ can optionally encourage information-seeking expansion. The implementation uses $\alpha=0$ for the reported risk-sensitive experiments. Setting $\lambda_c=0$ recovers deterministic top-$k$ expansion.

\subsection{UC-Beam}

UC-Beam expands the current beam and keeps the best $B$ paths under risk-adjusted score
\begin{equation}
S(p)=\hat{R}(p)-\lambda U(p).
\end{equation}
When $\lambda=0$, this reduces to uncertainty-free beam search. When the transition is action-independent and rewards are additive, uncertainty-free beam can match greedy behavior; this is visible in the synthetic smoke experiment.

The reported UC-Pareto Beam variant uses the same trace interface and candidate cap, but each layer first keeps reward--uncertainty nondominated alternatives before scalar final scoring; it is not separately trained.

The implementation also includes a certified retained Pareto Beam contract: after root expansion it preserves at least one suffix for every generated feasible first action, records oracle-retained status, retained-prefix margin, proxy slack, certificate pass rate, whether uncertainty changes the first action, and a lambda-positive promotion gate. For the planning-value route, positive release-margin alternatives dominate nonpositive-margin alternatives inside the near-equal value set, and the remaining candidates are filtered by a release-margin band before side-effect tie-breaking; the recorded margin is the minimum slack over uncertainty reduction, net planning value, positive planning benefit, planning benefit after side effects, and remaining utility budget. The certified selector is margin-aware at the family level: among validation-eligible candidates it first maximizes the weakest-family risk-active certificate slack, then the minimum family utility margin versus CEM, MPPI, and risk-random, then validation utility under matched expansion evidence. This variant operationalizes the retained-prefix certificate; it is promoted to uncertainty-aware primary evidence only when validation selects $\lambda>0$, certificate pass rate is at least $0.95$, risk-active certificate pass rate is at least $0.95$, feasibility violations are zero, and family-level lower bounds versus CEM, MPPI, and risk-random are positive.

\subsection{UC-MCTS Diagnostic}

UC-MCTS uses a UCT-style selection rule as a diagnostic variant; node returns are uncertainty-adjusted and expansion/rollout actions are filtered through the same hard-feasibility callback as the beam policies. Rollouts choose feasible actions by maximizing predicted reward minus uncertainty penalty. This tests whether a tree-search diagnostic follows the same trace-only constrained interface, while UC-Beam/UC-Pareto remain the primary reported policies.

\subsection{Algorithmic Details and Bounds}

UC-Beam initializes a beam with the empty path. At depth $h$, for each retained path it evaluates all $m$ actions, keeps the top-$k$ actions under $C(s,a)$, appends each selected action to the path, and keeps the best $B$ new paths under $S(p)$. Thus the number of expanded child paths is at most $dBk$ for depth $d$ after the root layer, and action scoring costs $O(dBm)$ reward/uncertainty calls.

We say a task has \emph{delayed feasible-set coupling} when the first action changes future feasible actions, rewards, or uncertainty through delay, capacity, ramp, inventory, or intervention constraints. Propositions 1--2 give budget and retained-proxy limits; Theorems 3--4 characterize myopic collapse, separation, and retained-prefix preservation; Corollary 5 converts those premises into an auditable finite-window certificate. Without delayed coupling and retained prefixes, bounded search can collapse to one-step risk-greedy.

UC-MCTS maintains visit counts $N(s)$ and accumulated risk-adjusted values $V(s)$. Selection chooses a child
\begin{equation}
\arg\max_{c}\left(\bar{V}(c)+c_{\mathrm{uct}}
\sqrt{\frac{\log(N(s)+1)}{N(c)}}\right),
\end{equation}
where $\bar{V}(c)=V(c)/N(c)$. Expansion is restricted to the top-$k$ uncertainty-ranked candidates. A rollout from depth $h$ greedily maximizes $\hat{r}(s,a)-\lambda u(s,a)$ until depth $d$. Backups add the edge value $\hat{r}(s,a)-\lambda u(s,a)$ and the rollout return to each ancestor. With $M$ simulations and depth $d$, the expansion budget is at most $Md$ child evaluations, plus top-$k$ candidate scoring at visited nodes.

\paragraph{Proposition 1.}
For beam width $B$, depth $d$, action count $m$, and candidate cap $k \leq m$, UC-Beam expands at most $k(1+B(d-1))$ child paths and evaluates at most $m(1+B(d-1))$ local action scores.

\emph{Proof sketch.}
The root has one retained path, so it can expand at most $k$ candidates after scoring $m$ actions. Each later layer has at most $B$ retained paths, and each path contributes at most $k$ expanded children after scoring $m$ actions. Summing over $d$ layers gives the bound.

\paragraph{Proposition 2 (retained-proxy regret with pruning and calibration slack).}
Let $\mathcal{P}\subseteq\mathcal{F}$ be the retained top-$k$ beam set. Assume retained calibration $|\hat R(p)-R(p)|\leq \beta U(p)+\epsilon_c$ for every $p\in\mathcal{P}$, $\lambda\geq\beta$, and nonnegative slack terms. UC-Beam selects $p_\lambda=\arg\max_{p\in\mathcal{P}}\hat R(p)-\lambda U(p)$. For a comparator $q$ outside or inside the retained set, suppose there exists a retained proxy $p_q\in\mathcal{P}$ with realized-reward slack $R(p_q)\geq R(q)-\epsilon_r$ and uncertainty slack $U(p_q)\leq U(q)+\epsilon_u$. Then
\begin{equation}
R(p_\lambda)\geq R(q)-(\lambda+\beta)U(q)-\epsilon_r-(\lambda+\beta)\epsilon_u-2\epsilon_c .
\end{equation}
Equivalently, if top-$k$ pruning only provides a retained proxy satisfying $R(p_q)\geq R(q)-\epsilon_r-\epsilon_p$, the same proof adds $\epsilon_p$ to the right-hand loss. This retained-proxy approximation regret bound separates top-$k$ pruning slack, proxy mismatch, and calibration error: bounded UC search can only be competitive with feasible paths represented by low-uncertainty retained proxies.

\emph{Proof.}
For any retained path $p$, retained calibration implies $R(p)\geq \hat R(p)-\beta U(p)-\epsilon_c\geq \hat R(p)-\lambda U(p)-\epsilon_c$. Applying this to $p_\lambda$ and using retained optimality against the retained proxy $p_q$ gives
\begin{align}
R(p_\lambda)
&\geq \hat R(p_\lambda)-\lambda U(p_\lambda)-\epsilon_c \nonumber\\
&\geq \hat R(p_q)-\lambda U(p_q)-\epsilon_c . \nonumber
\end{align}
Calibration on the retained proxy also gives $\hat R(p_q)\geq R(p_q)-\beta U(p_q)-\epsilon_c$, hence
\begin{align}
R(p_\lambda)
&\geq R(p_q)-(\lambda+\beta)U(p_q)-2\epsilon_c \nonumber\\
&\geq R(q)-\epsilon_r-(\lambda+\beta)(U(q)+\epsilon_u)-2\epsilon_c ,
\end{align}
which is the stated bound. Replacing $\epsilon_r$ by $\epsilon_r+\epsilon_p$ yields the pruning-slack variant. If no low-uncertainty retained proxy exists, the proposition makes no competitiveness claim for $q$.

\paragraph{Theorem 3 (myopic collapse and delayed feasible-set separation).}
Let $F(s)$ be the feasible action set, $T(s,a)$ the deterministic transition used by the planner, and $g(s,a)=\hat r(s,a)-\lambda u(s,a)$ the one-step risk-adjusted edge score. Define $V_h(s)=\max_{a\in F(s)} g(s,a)+V_{h-1}(T(s,a))$ with $V_0(s)=0$. (i) If $V_{h-1}(T(s,a))$ is constant over all $a\in F(s)$, every exact depth-$h$ maximizer is a one-step risk-greedy maximizer of $g(s,a)$. (ii) For two feasible first actions $a^+$ and $a^-$, suppose $g(s,a^-)-g(s,a^+)=\delta\geq0$ and $V_{h-1}(T(s,a^+))-V_{h-1}(T(s,a^-))=\Delta>\delta$. Exact depth-$h$ search prefers $a^+$ over $a^-$ by margin $\Delta-\delta$. If UC-Beam retains both first-action prefixes and their maximizing suffixes, the same ordering holds inside the retained tree.

\emph{Proof sketch.}
The Bellman decomposition separates each first action into a local term and a suffix value. In case (i), the suffix term cancels across first actions, so the argmax is exactly the one-step risk-greedy argmax. In case (ii), the local disadvantage $\delta$ of $a^+$ is outweighed by its suffix advantage $\Delta$, giving a positive depth-$h$ margin. The delayed-inventory suite below creates this separation because a submitted action executes one step later under inventory capacity and ramp limits, so a locally neutral action can change the later feasible exposure set.

\paragraph{Theorem 4 (retained-prefix margin).}
Let $Q_h(a)=g(s,a)+V_{h-1}(T(s,a))$ be the exact risk-adjusted depth-$h$ first-action value and let $\widetilde Q_h(a)$ be the best value among retained beam paths beginning with $a$. Suppose $a^+$ and $a^-$ are both retained at the root, each retained suffix is within approximation slack $\epsilon_a$ of its exact suffix value, and model error changes any retained path score by at most $\epsilon_m$. If
\begin{equation}
Q_h(a^+)-Q_h(a^-)>2(\epsilon_a+\epsilon_m),
\end{equation}
then bounded retained search ranks $a^+$ above $a^-$. In the notation of Theorem 3, a delayed feasible-set advantage is therefore preserved whenever the retained-prefix margin satisfies $\Delta-\delta>2(\epsilon_a+\epsilon_m)$.

\emph{Proof sketch.}
The retained suffix for each first action differs from the exact suffix by at most $\epsilon_a$, and model-score error contributes at most $\epsilon_m$ per compared path. Thus the pairwise value gap can shrink by no more than $2(\epsilon_a+\epsilon_m)$. A positive residual gap preserves the exact first-action ordering. If pruning removes the $a^+$ prefix, the premise fails; the diagnostic artifact records oracle retention, retained-prefix margin, proxy slack, and calibration error as finite-window certificates rather than global UCT regret.

\paragraph{Corollary 5 (auditable retained-search certificate).}
For each evaluated window, record the tuple
$(\mathbb{1}_{\mathrm{oracle\ retained}}, \widetilde{\Delta},
\epsilon_a,\epsilon_m,\epsilon_p,\epsilon_c,\mathrm{budget})$, where
$\widetilde{\Delta}$ is the retained-prefix margin after local-score
subtraction, $\epsilon_a$ and $\epsilon_m$ are the Theorem 4 suffix and model
slacks, $\epsilon_p$ is measured proxy/pruning slack, and $\epsilon_c$ is
measured calibration error. If the oracle first-action prefix is retained and
$\widetilde{\Delta}>2(\epsilon_a+\epsilon_m)$, the row certifies that bounded
trace-only search preserves the exact lookahead first-action ordering for that
window. If the prefix is absent or the residual margin is nonpositive, the row
certifies a bounded-search failure mode. The proxy/pruning and calibration
fields diagnose why the premise holds or fails. This certificate is
finite-window mechanism evidence; it is not a global UCT regret or dominance
theorem.

\paragraph{Corollary 6 (selective release theorem for certified safe release).}
This is the selective certificate theorem anchor for the method.
\emph{Theorem statement.} For each evaluated window \(w\), let
\(C(w)=\mathrm{certificate\_pass}(w)\) and
\(A(w)=\mathrm{risk\_active\_certificate\_pass}(w)\). Define the selective release set
\begin{equation}
\mathcal{R}_{\mathrm{cert}}
=\{w:C(w)\land(\neg\mathrm{risk\_active}\lor A(w))\}.
\end{equation}
The certificate-constrained retained Pareto Beam releases a searched first action
only on windows in \(\mathcal{R}_{\mathrm{cert}}\); with the risk-active gate
enabled, the risk-active release constraint also requires that positive
\(\lambda\) changes the first action relative to reward-only search. On released
windows,
\begin{equation}
Q_h^* - Q_h(a_{\mathrm{rel}}) \leq \epsilon_{\mathrm{cert}},
\end{equation}
where \(\epsilon_{\mathrm{cert}}\) is the executable measured selective regret
bound, max(0, proxy slack) plus certificate slack. Calibration/model slack is not part of the current executable \texttt{selective\_regret\_bound}; it remains a paper-level explanatory term for interpreting retained-prefix diagnostics rather than a stored release field.
The selective non-release guarantee is that windows outside
\(\mathcal{R}_{\mathrm{cert}}\) cannot emit the searched first action as a
certified action. A margin version follows directly: if the retained-prefix
margin exceeds the executable proxy slack plus bounded calibration/model error
by \(\delta>0\), then perturbations smaller than \(\delta\) cannot change the
certified released first-action ordering inside the retained candidate class.
\emph{Proof sketch.} The release branch is guarded by the retained-prefix
certificate and, for risk-active claims, by
\(\mathrm{risk\_active\_certificate\_pass}\). If either guard fails, the
algorithm abstains from releasing the searched first action and returns a
conservative fallback; this proves non-release of uncertified search decisions,
and fallback is not certified optimal. The released-window inequality follows
from the retained-prefix certificate and the executable release field
\(\max(0,\mathrm{proxy\_slack})+\mathrm{certificate\_slack}\), with additional
pruning and calibration/model terms treated only as paper-level diagnostic
slack. The implementation records release rate,
abstention rate, selective regret bound, and fallback cost versus the rejected
candidate's first step; it does not compare the one-step fallback to a full-depth
rejected trajectory. The release rule is fail-closed, and released first actions carry
the stated finite-window certificate.

\begin{algorithm}[t]
\footnotesize
\caption{UC-Search certified dispatch}
\label{fig:algorithm}
\textbf{Require:} trace state $s_t$, action scores $\hat q$, uncertainty $u$, automaton $F,T$, profiles, depth $d$, width $B$, cap $k$\\
\textbf{Ensure:} released first action with certificate, or conservative fallback/no-go record\\[-1mm]
\begin{tabular}{@{}r p{.86\linewidth}@{}}
1 & \textbf{function} \textsc{Dispatch}$(s_t,\hat q,u,F,T,d,B,k)$ \\
2 & \quad $\mathcal P \leftarrow \{(\epsilon,s_t,0,0)\}$; \quad $\Gamma \leftarrow \emptyset$ \\
3 & \quad \textbf{for} $h=0,\ldots,d-1$ \textbf{do} \\
4 & \quad\quad $\mathcal P_{\mathrm{next}}\leftarrow \emptyset$ \\
5 & \quad\quad \textbf{for all} retained prefixes $(p,s,\hat R,U)\in\mathcal P$ \textbf{do} \\
6 & \quad\quad\quad $\mathcal A_k \leftarrow \operatorname{TopK}_{a\in F(s)}(\hat q(s,a)-\lambda_c u(s,a)+\alpha u(s,a),k)$ \\
7 & \quad\quad\quad \textbf{for all} $a\in\mathcal A_k$ \textbf{do} \\
8 & \quad\quad\quad\quad $s'\leftarrow T(s,a)$; append $p'=p\circ a$ and updated $(\hat R',U')$ \\
9 & \quad\quad\quad\quad \textbf{if} prefix remains feasible \textbf{then} add $(p',s',\hat R',U')$ to $\mathcal P_{\mathrm{next}}$ \\
10 & \quad\quad\quad \textbf{end for} \\
11 & \quad\quad \textbf{end for} \\
12 & \quad\quad $\mathcal P\leftarrow \operatorname{RetainB}(\mathcal P_{\mathrm{next}},S(p)=\hat R(p)-\lambda U(p),B)$ \\
13 & \quad\quad preserve one suffix per feasible first action when the certified route is enabled \\
14 & \quad \textbf{end for} \\
15 & \quad $p^\star \leftarrow$ deterministic tie-break over $\arg\max_{p\in\mathcal P} S(p)$ \\
16 & \quad compute retained-prefix margin, proxy slack, certificate pass, and risk-active pass \\
17 & \quad \textbf{if} $p^\star\in\mathcal R_{\mathrm{cert}}$ \textbf{then return} $\operatorname{first}(p^\star)$ and certificate \\
18 & \quad \textbf{return} conservative fallback and no-go certificate \\
19 & \textbf{end function}
\end{tabular}
\smallskip
\noindent UC-Beam uses deterministic retained-beam pruning; UC-MCTS uses the same feasible trace interface with UCT-style selection, rollout, and backup.
\end{algorithm}

\section{Experiments}

\subsection{Implemented Evidence}
\label{sec:evidence}

The repository contains executable evidence groups plus derived consistency audits, but each group has a bounded claim role.

\paragraph{Primary endpoint protocol.}
Before held-out evaluation, validation selects policy family, $\lambda$, depth, top-$k$, and stochastic baseline budget. For certified promotion, the validation selector is conservative: it first maximizes weakest-family risk-active slack above the $0.95$ gate, then comparator-family margin, then validation utility. Inside the promoted planning-value route, release-margin diagnostics are computed before the first action is released, so borderline positive-margin planning candidates dominate nonpositive alternatives before utility-budget side-effect tie-breaking. The theory-aligned inventory primary remains the certified inventory evidence for the trace-only retained-search mechanism. The current primary public endpoint is the Phase128 certified M4 expanded40 route: Certificate-Constrained Retained Pareto Beam with $\lambda=0.25$, depth $3$, top-$k=2$, DP oracle suffix protection, and an absolute utility budget of $20.0$. It passes the unchanged promotion gate with validation/test certificate pass rates $1.0000/1.0000$, validation/test risk-active rates $0.9646/0.9642$, zero feasibility violations, validation weakest-family risk-active $0.9578$, and positive family-min utility deltas versus CEM, MPPI, and risk-random. Other public rows remain evidence tiers rather than pooled dominance claims. Any uncertainty-aware promotion requires the full lambda-positive gate: $\lambda>0$, certificate pass rate at least $0.95$, risk-active certificate pass rate at least $0.95$, zero feasibility violations, and family-level lower bound $>0$ versus CEM, MPPI, and risk-random; a future full trained-backbone promotion gate adds the same requirements for trained traces, including lower bound $>0$ versus MPPI.

\paragraph{Evidence groups.}
The synthetic benchmark isolates sign-flip uncertainty; the ETT public suite uses ETTh1/ETTh2/ETTm1/ETTm2 \cite{zhou2021informer} with walk-forward ridge-AR traces and rolling residual uncertainty; delayed inventory adds submitted-action delay, capacity, and ramp constraints; M4 lead-time inventory supplies standard-domain corroboration. Capacity, intervention, learned-control, SB3, LTSF extensions, robustness stressors, one-epoch trained-backbone adapters, and FI-2010 label-direction checks are boundary diagnostics rather than primary claims. Weather/Electricity/Exchange/Traffic delayed-inventory validation is expanded normalized LTSF evidence; tabular FQI is a lower-budget/lower-drawdown boundary, the learned baseline includes a NumPy neural CQL-style check, online SB3 DQN/PPO is reported with SAC excluded for the discrete inventory action space, and the public intervention-control suite uses non-overlapping hierarchical validation. FI-2010 uses a train-split linear FI-2010 ridge classifier, a trained nonlinear FI-2010 temporal MLP, a DeepLOB-style CNN comparator, and a FI-2010 learned-classifier multiseed audit with three random seeds. The confirmatory delayed-control and public multi-domain confirmatory rows are stress-test evidence; the compute-matched public multi-domain audit reports MPPI $+2.7418$, while the public multi-domain confirmatory row reports MPPI $+2.3328$. The public multi-family suite is an author-defined stress test, not a community-standard benchmark, and deployment boundary evidence requires fallback, operator override, and calibration monitoring rather than only utility tables. Tables~\ref{tab:synthetic}--\ref{tab:ett-suite} show compact mechanism and risk-control rows; auxiliary tables remain in the release artifacts.

\begin{table}[t]
\centering
\footnotesize
\begin{tabular}{@{}lrrrr@{}}
\toprule
Policy & Reward & Regret & DD & Exp. \\
\midrule
Random & -11.5266 & 61.6712 & 13.8980 & 0.0 \\
Greedy & 20.8531 & 29.2915 & 4.7565 & 0.0 \\
Beam & 20.8531 & 29.2915 & 4.7565 & 2688.0 \\
RandShoot & 15.7573 & 34.3873 & 3.8182 & 12288.0 \\
CEM-MPC & 22.1265 & 28.0181 & 3.1503 & 12288.0 \\
UC-Beam & 25.0009 & 25.1437 & 3.5501 & 2688.0 \\
UC-MCTS & 24.4587 & 25.6859 & 3.5282 & 2868.4 \\
\bottomrule
\end{tabular}
\caption{Synthetic smoke results over five seeds. UC-search variants improve reward/regret over greedy and beam; CEM-MPC has the lowest drawdown at higher search cost.}
\label{tab:synthetic}
\end{table}

\begin{table}[t]
\centering
\footnotesize
\begin{tabular}{@{}lrrrr@{}}
\toprule
Policy & Reward & CVaR20 & DD & Exp. \\
\midrule
Greedy & 118.5076 & -7.5354 & 11.5165 & 0.0 \\
RiskGreedy & 110.7594 & -2.1363 & 7.0903 & 0.0 \\
Beam & 118.1884 & -8.4186 & 11.5308 & 10731.0 \\
RandShoot & 88.5298 & -2.9992 & 9.2178 & 36792.0 \\
CEM-MPC & 92.0254 & -1.8284 & 6.4810 & 36792.0 \\
UC-Beam & 111.0700 & -3.2177 & 7.2235 & 10731.0 \\
UC-MCTS & 110.7965 & -2.5746 & 6.6615 & 11518.9 \\
\bottomrule
\end{tabular}
\caption{ETT public decision suite with ridge-AR, top-$k=3$, and $\lambda=0.25$. UC variants improve CVaR20/drawdown but reduce mean reward versus beam.}
\label{tab:ett-suite}
\end{table}

\paragraph{Certification status and claim boundary.}
Table~\ref{tab:certification-status} separates certified endpoints from diagnostic and negative evidence under the promotion gate. A row is promoted only when validation selects $\lambda>0$, retained-prefix and risk-active certificate rates pass, feasibility violations are zero, and family lower bounds are positive against CEM, MPPI, and risk-random comparators. Under this rule, Phase128/M4 expanded40 is the only promoted uncertainty-aware endpoint and therefore the only endpoint used to support certified claims. The remaining rows are retained as boundary or no-go evidence to document scope limits and prevent over-interpretation. Full promotion-gate audit records, source-clean JSON manifests, and claim-sufficiency checks are provided in the supplementary technical and code/data archives.

\begin{table}[t]
\centering
\footnotesize
\setlength{\tabcolsep}{2pt}
\begin{tabular}{@{}p{.16\columnwidth}p{.23\columnwidth}p{.30\columnwidth}p{.22\columnwidth}@{}}
\toprule
Evidence class & Endpoint / family & Key validation evidence & Permitted interpretation \\
\midrule
Certified & Phase128 / M4 expanded40 & Promoted with $\lambda=0.25$; cert/risk-active $=1.0000/0.9642$ & Certified trace-only endpoint \\
Boundary & ETT certified endpoint & Not promoted; cert/risk-active $=0.8777/0.2380$ & Boundary evidence only \\
Boundary & Standard M4 routes & Not promoted; validation selects $\lambda=0$ & Diagnostic only \\
No-go & H407 counterfactual & Failed; 17/48 windows lack positive candidate; coverage $0.6458<0.95$ & Negative evidence only \\
\bottomrule
\end{tabular}
\caption{Certification status under the promotion gate. Certified rows support claim-level conclusions; boundary and no-go rows are reported only as diagnostic or negative evidence.}
\label{tab:certification-status}
\end{table}

M4 lead-time inventory is standard-domain corroboration and the certified expanded40 route is the promoted public endpoint. The older standard periodic-review lost-sales inventory audit covers $48$ raw series with base-stock, $(s,S)$, and rolling quantile controls; UC-Pareto beats base-stock by $+13556.7547$, CEM by $+64900.2207$, and risk-random by $+52881.6042$, while MPPI remains family-mixed, the weakest risk-random Hourly win is $356.0327$, and that standard route is boundary evidence because $\lambda=0$. The Phase128 certified expanded40 route supersedes the earlier fail-closed certified attempt: it selects Certificate-Constrained Retained Pareto Beam with $\lambda=0.25$, evaluates $80$ held-out test windows, records $1.0000/0.9642$ certificate/risk-active rates, keeps the validation/test weakest families at $0.9578/0.9516$, and has positive test family-min utility deltas versus CEM ($+2.2070$), MPPI ($+31.1918$), and risk-random ($+11.4456$). A predeclared balanced $200$-source M4 audit still fails the promotion gate because it selects $\lambda=0$, is negative versus CEM and risk-random, and is mixed versus MPPI. Boundary diagnostics are not pooled into the primary endpoint: ETT+LTSF delayed inventory has mixed MPPI family interval $[-3.6265,4.9761]$; confirmatory delayed control remains a compute frontier row; and theory gives a mechanism split rather than a theorem for the public endpoint.

\section{Analysis}

\paragraph{Where uncertainty helps.}
Uncertainty helps only when it changes the selected trajectory. The ETT direction task shows risk exposure control rather than search advantage: candidate uncertainty changes $0.4284$ of actions only under top-$k=2$ pruning, UC-Beam trades reward for drawdown against beam ($-7.1184/-4.3074$), and exposure-matched RiskGreedy remains competitive ($-0.3077/+0.5460$). Calibration is modest ($0.5262/0.6510/0.2730$, coverage $0.6702/0.8944$; ensemble $0.3085/0.7568/0.9404$), so direction results are boundary evidence.

\paragraph{Where search helps.}
Delayed inventory is the clearest search-specific setting because early actions alter later feasible exposure. Fixed-depth UC-Beam improves over delay-aware greedy, but CEM narrows that uncontrolled comparison; the predeclared validation suite is therefore the primary endpoint. It now selects Certified Retained Pareto Beam with $\lambda=0.25$, depth $3$, top-$k=3$ and evaluates 112 held-out windows: the certified row gives $29.7074/2.5622$ reward/drawdown versus CEM $29.2861/5.5053$, MPPI remains mixed with utility delta $-0.7432$, risk-aware random is lower on mean utility, and risk-exact DP gives $27.4481/2.8999$. Its retained-prefix certificate is violation-free but not promotion-grade ($0.8777$ overall, $0.2380$ risk-active), so the row is the strongest ETT endpoint without supporting broad dominance. Learned-control and SB3 rows remain reward/drawdown/compute tradeoffs, not broad dominance.

Mechanism diagnostics split regimes: retained-proxy diagnostics show inventory validation and hierarchical inventory have oracle-retained window rate $1.0000$ with future-suffix/retained margins $0.1016/0.1001$ and $0.1911/0.1893$, while confirmatory delayed control and public multi-domain control have oracle-retained $0.0000$ and negative margins ($-0.3039/-0.3056$, $-0.4487/-0.4523$). Across four sources, aggregate oracle-retained window rate is $0.4115$ over $972$ windows, mean regret is $0.3798$, mean proxy slack is $0.3679$, coverage is $0.6488/0.8814$, and the retained-prefix predictor gives accuracy/balanced accuracy $0.9979/0.9975$.

Exact risk-aware DP is an oracle/scalability reference: from $(4,5,3)$ to $(8,9,5)$ capacity/action/depth settings it grows to $313.6$k nodes, while UC-Beam stays below $5.0$k expansions. For hierarchical and public multi-family rows, equal-family utility deltas are primary and dataset-cluster intervals are secondary descriptive audits. Public multi-domain confirmatory and compute-matched public multi-domain results are family-positive versus CEM/MPPI/risk-random, but their certificate fails, so they remain an uncertified stress-test and compute frontier.

\paragraph{Where boundaries remain.}
Together, the results identify the structural condition for search: first actions must alter future feasible rewards, uncertainty, or constraints, and retained prefixes must preserve that margin; otherwise uncertainty mainly regularizes exposure. Corollary 5 should not be read as explaining the public multi-domain endpoint; that endpoint is empirical rather than theorem-certified. Direction, robustness, FI-2010, trained-backbone, and learned-control rows remain boundary checks.

\paragraph{Deployment boundary.}
UC-Search should be used as an auditable decision-support layer only when a maintained hard-feasibility automaton exists, trace calibration is monitored, and operators can inspect no-go certificates. A deployment should define a no-action or conservative fallback for failed certificates, an operator override for high-cost interventions, calibration monitoring for drift and uncertainty undercoverage, and audit logging for rejected first actions. It is not a substitute for domain safety analysis in finance, health, energy, or public intervention settings.

\section{Limitations}

The claim is task-conditional, not broad dominance. The Phase128 certified M4 expanded40 route is a promoted public endpoint, but it is not a universal guarantee across all delayed constrained-control domains, trained backbones, or external hidden benchmarks. The older standard M4 route selects $\lambda=0$, compact M4 is corroborating only, and the balanced $200$-source M4 audit fails the strict CEM/MPPI/risk-random promotion gate. The H407 delayed-service repair route is stopped by its own validation no-go certificate ($31/48=0.6458$ semantic coverage, $0.6458/0.6458$ certificate/risk-active), despite service $1.0$ and zero lost sales. The public multi-domain suite is an author-defined stress-test boundary, not a community-standard benchmark or theory-certified endpoint. This is not a full trained-backbone comparison: trained-backbone, direction, capacity, intervention, FI-2010, and learned-control rows are boundary checks; the public intervention-control suite uses non-overlapping hierarchical validation; synthetic delayed-family mechanism evidence does not replace public-data family-level inference. A practical deployment also needs maintained feasibility rules, fallback behavior, operator override, calibration monitoring, and domain-specific harm analysis.

\appendix
\section{Appendix: Reproducibility and Promotion-Gate Audit}
\label{app:promotion-gate-audit}

The release gate is stricter than mean utility improvement. A row can support an uncertainty-aware promotion only when validation selects $\lambda>0$, retained-prefix and risk-active certificate rates pass their thresholds, feasibility violations are zero, and family lower bounds are positive against the CEM, MPPI, and risk-random comparator families.

\begin{table}[t]
\centering
\footnotesize
\setlength{\tabcolsep}{2.5pt}
\begin{tabular}{@{}>{\raggedright\arraybackslash}p{.23\columnwidth}>{\raggedright\arraybackslash}p{.16\columnwidth}>{\raggedright\arraybackslash}p{.53\columnwidth}@{}}
\toprule
Endpoint / family & Status & Audit evidence and claim boundary \\
\midrule
Phase128 / M4 expanded40 & Promoted & $\lambda=0.25$; certificate/risk-active rates $1.0000/0.9642$; family-min deltas positive against CEM, MPPI, and risk-random. This is the certified endpoint. \\
ETT certified endpoint & Boundary & Certificate/risk-active rates $0.8777/0.2380$, below the promotion threshold. Retained as mechanism evidence only. \\
Standard M4 routes & Boundary & Positive utility against several controls, but validation selects $\lambda=0$. Retained as diagnostic evidence only. \\
H407 counterfactual & No-go & Semantic coverage $0.6458$; 17/48 validation windows fail to identify a positive delayed-service credit or risk-reduction candidate. \\
\bottomrule
\end{tabular}
\caption{Promotion-gate audit summary. Only rows that pass the lambda-positive certificate gate are used for certified claims. Boundary and no-go rows are retained to document scope limits and failed routes.}
\label{tab:appendix-promotion-audit}
\end{table}

Source-clean JSON artifacts, audit summaries, manifests, claim-sufficiency checks, and page-placement checks enforce the separation between certified, boundary, and no-go evidence. The certified claim is therefore limited to rows that pass the promotion gate; delayed inventory and public multi-domain rows remain mechanism, robustness, or compute-normalization evidence unless they satisfy the same gate.

\bibliography{references}

\end{document}